\documentclass[journal]{IEEEtran}
\usepackage{cite}
\usepackage{graphicx, amsfonts, amsmath, amssymb}
\usepackage{float}
\usepackage{booktabs}
\usepackage{algorithm}
\usepackage{algpseudocode}
\ifCLASSINFOpdf

\else

\fi

\begin{document}

\title{Studying the Effects of Self-Attention on SAR Automatic Target Recognition}

\author{Jacob Fein-Ashley$^{*}$,
    Rajgopal Kannan$^{\dagger}$,
    Viktor Prasanna$^{*}$
}

\author{
    \IEEEauthorblockN{Jacob Fein-Ashley\IEEEauthorrefmark{1}, Rajgopal Kannan\IEEEauthorrefmark{2}, Viktor Prasanna\IEEEauthorrefmark{1}}\\
    \IEEEauthorblockA{\IEEEauthorrefmark{1}University of Southern California}
    \IEEEauthorblockA{\IEEEauthorrefmark{2}DEVCOM Army Research Office}
}

\markboth{}%
{Fein-Ashley \MakeLowercase{\textit{et al.}}: Studying the Effects of Self-Attention on SAR ATR}

\maketitle

\begin{abstract}
Attention mechanisms are critically important in the advancement of synthetic aperture radar (SAR) automatic target recognition (ATR) systems. Traditional SAR ATR models often struggle with the noisy nature of the SAR data, frequently learning from background noise rather than the most relevant image features. Attention mechanisms address this limitation by focusing on crucial image components, such as the shadows and small parts of a vehicle, which are crucial for accurate target classification. By dynamically prioritizing these significant features, attention-based models can efficiently characterize the entire image with a few pixels, thus enhancing recognition performance. This capability allows for the discrimination of targets from background clutter, leading to more practical and robust SAR ATR models. We show that attention modules increase top-1 accuracy, improve input robustness, and are qualitatively more explainable on the MSTAR dataset.
\end{abstract}

\begin{IEEEkeywords}
SAR ATR, Synthetic Aperture Radar, Attention, Deep Learning
\end{IEEEkeywords}

\IEEEpeerreviewmaketitle

\section{Introduction}

In the realm of SAR ATR, the ability to accurately and efficiently identify targets is paramount for surveillance, reconnaissance, and remote sensing applications. Traditional SAR ATR models often face challenges due to the high-dimensional and noisy nature of SAR data, leading them to inadvertently learn from background noise rather than the most critical features of the image. However, recent advances in attention mechanisms~\cite{ecanet,cbam,squeezeexcitation} have revolutionized the field of image classification by allowing models to focus on the most prominent parts of the image, such as the shadows and distinct components of a vehicle for SAR ATR applications. These attention-based approaches can characterize the entire image with just a few pixels, significantly improving the practical performance of the model. By focusing on these crucial aspects, attention mechanisms enhance the ability of SAR ATR systems to differentiate between targets and background clutter, paving the way for more accurate, explainable, and reliable target recognition in diverse and challenging environments.

In SAR ATR, a chosen deep learning model will invariably learn from the backscatter associated with the SAR image~\cite{noncausality}. We find that by applying attention modules such as~\cite{ecanet,squeezeexcitation,cbam}, the model can learn more about the essential parts of the image. In addition, the literature related to SAR ATR has found the attention mechanism useful~\cite{attentionsaratr,physicsattention, saratratt}, achieving state-of-the-art accuracy with their chosen datasets.

\begin{figure}[H]
    \centering
    \includegraphics[width=\columnwidth]{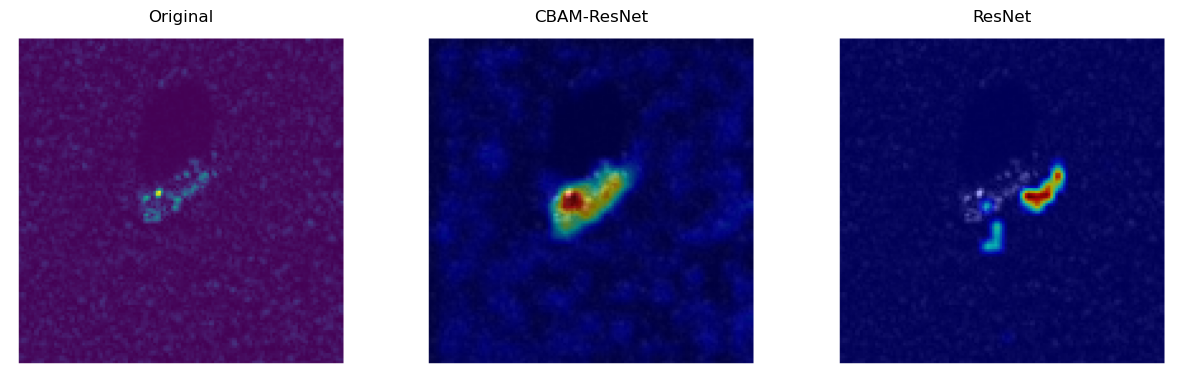}
    \caption{Original Image vs. Grad-CAM~\cite{gradcam} CBAM Attention Model~\cite{squeezeexcitation} with a ResNet-18~\cite{resnet} Backbone vs. ResNet-18}
    \label{fig:comparison}
\end{figure}

\textbf{Visual Attention. }
The visual attention mechanism imitates human cognitive awareness of a particular information, focusing on critical details to focus more on the essential elements of the data. Attention allows models to dynamically focus on the most relevant parts of the input data while processing it. It mimics the human cognitive process of selectively concentrating on certain input aspects while ignoring others. This selective focus enables the model to handle complex tasks more effectively by prioritizing critical information for decision-making~\cite{visualattention}. Refer to Fig~\ref{fig:comparison}; we see that the ResNet-18~\cite{resnet} model with an added CBAM attention module can focus on the vehicle rather than learning from seemingly random parts of the image.

In this paper, we study the effects of self-attention mechanisms on SAR ATR and how self-attention can create more \textbf{accurate}, \textbf{robust}, and \textbf{explainable} algorithms. We use the MSTAR data set (Moving and Stationary Target Acquisition and Recognition) to qualitatively and quantitatively evaluate the decisions of a ResNet-18 model. Our experimental setup includes a ResNet-18 backbone model with state-of-the-art Convolutional Block Attention Module (CBAM)~\cite{cbam}, SENet~\cite{squeezeexcitation}, and ECANet~\cite{ecanet} attention mechanisms placed in the ResNet-18 backbone. We use a ResNet-18 model as the standard backbone to test each attention module, as it performs well in the SAR-ATR setting~\cite{benchmark} and is convolutional-based, allowing all self-attention mechanisms to be ``plugged'' into the model. The computational performance of each of these mechanisms varies. We evaluate their effectiveness using a standard quantitative accuracy measure (top-1 accuracy with 3 tests, refer to Table~\ref{tab:mstar}), robustness through an input perturbation accuracy measure (refer to Table~\ref{tab:robustness}, and qualitative heat map visualizations using Grad-CAM~\cite{gradcam}. In Section II, we describe details of the various attention mechanisms, while in Section 3, we obtain experimental results using self-attention on a ResNet-18 backbone.

\section{Self-Attention Methods}
The self-attention mechanism is a prediction technique that allows the model to focus on different parts of the input sequence.  Visual attention mechanisms usually emphasize the essential parts of the data while reducing the importance of less critical parts. This process draws inspiration from the human cognitive system~\cite{itti, att}. Pioneering work in attention~\cite{cbam, squeezeexcitation, ran} uses CNN as the backbone, allowing their attention modules to be placed on any convolutional-based model. These works use attention maps to scale feature mappings in CNNs. Recent attention mechanisms have tried to reduce the complexity of such attention models~\cite{ecanet} while improving or maintaining accuracy in image classification tasks. Furthermore, ~\cite{integration} shows that convolution and attention can seamlessly integrate, as they have similar operations. This study focuses on self-attention mechanisms used in CNNs. From studies of~\cite{integration}, it is clear that self-attention can be seamlessly integrated into CNNs, as they are nearly the same operation. 

The importance of attention can be qualitatively visualized through the Grad-CAM process~\cite{gradcam}, which stands for Gradient-weighted Class Activation Mapping. It is a technique used in machine learning, specifically in deep learning and convolutional neural networks (CNNs), to provide visual explanations for the decisions made by these models. It helps in understanding which regions of an input image are essential for the predictions made by the network. We use Grad-CAM in our studies to visualize the important pixels in the SAR images that contribute the most to a ResNet model's decision.

Recent advances in attention mechanisms such as~\cite{vit, cbam, ecanet,squeezeexcitation} have allowed convolutional neural networks (CNNs) and transformer architectures to focus on crucial parts of input data. In CNN-based models, self-attention mechanisms can allow for more interpretability, robustness, and accuracy in some cases. In SAR ATR, a model must be explainable to those who use it. By plotting the visual explanation of Grad-CAM for a sample SAR image, it is clear that in Fig.~\ref{fig:comparison}, the ResNet-based model does not focus on the vehicle and crucial parts of the image, while the CBAM-ResNet-18 model can capture essential parts of the image, such as parts of the vehicle and some of the shadow region cast from the vehicle.

The importance of self-attention in the medical imaging community is highlighted by~\cite{studyingtheeffects}. Furthermore, \cite{studyingtheeffects} emphasize the importance of self-attention in their field, discussing that self-attention mechanisms may be more in line with how a clinician would classify an image. Medical image classification is similar to SAR ATR in that small parts or characteristics of an image can significantly affect a model's decision.

\subsection{Formulation of Self-Attention}
\label{sec:formulation}
Consider a CNN with $L$ layers, where the output of the $l$-th layer is denoted as $X^{(l)} \in \mathbb{R}^{C \times H \times W}$, where $C$ is the number of channels and $H$ and $W$ are the height and width of the feature map, respectively. The self-attention mechanism is applied to the input $X^{(l)}$ to produce the output $Y^{(l)} \in \mathbb{R}^{C \times H \times W}$. The self-attention mechanism is defined as:
\begin{equation*}
    Y^{(l)} = X^{(l)} + \text{SA}(X^{(l)}),
\end{equation*}
where $\text{SA}(\cdot)$ is the self-attention mechanism. Thus, the output of the $l$-th layer is the sum of the input and the self-attention mechanism applied to the input. Furthermore, our studies use the self-attention mechanism proposed in~\cite{squeezeexcitation, cbam, ecanet}, recent state-of-the-art advances in self-attention. We explain these mechanisms in the following sections.

\subsection{CBAM Model}
The Convolutional Block Attention Module (CBAM)~\cite{cbam} is a lightweight yet effective attention module for feedforward convolutional neural networks. To refine feature maps, it sequentially infers attention maps along two separate dimensions, channel and spatial.

\subsubsection{Channel Attention Module}
The Channel Attention Module (CAM) generates an attention map along the channel dimension. It aggregates spatial information using average and max-pooling operations and then produces the attention map using a shared network.

\begin{align*}
    \mathbf{F}_{avg}^c &\in \mathbb{R}^{C \times 1 \times 1} = \text{AvgPool}(\mathbf{F}) \in \mathbb{R}^{C \times H \times W} \\
    \mathbf{F}_{max}^c &\in \mathbb{R}^{C \times 1 \times 1} = \text{MaxPool}(\mathbf{F}) \in \mathbb{R}^{C \times H \times W}
\end{align*}

The shared network comprises a multilayer perceptron (MLP) with one hidden layer. The two aggregated features are passed through the MLP to produce the channel attention map.

\begin{align*}
    \mathbf{M}_c(\mathbf{F}) &\in \mathbb{R}^{C \times 1 \times 1} = \sigma(\text{MLP}(\mathbf{F}_{avg}^c) + \text{MLP}(\mathbf{F}_{max}^c)) \\
    \mathbf{F}_c' &\in \mathbb{R}^{C \times H \times W} = \mathbf{M}_c(\mathbf{F}) \otimes \mathbf{F} \in \mathbb{R}^{C \times H \times W}
\end{align*}

Here, $\sigma$ denotes the sigmoid function, and $\otimes$ denotes element-wise multiplication.

\subsubsection{Spatial Attention Module}
The Spatial Attention Module (SAM) generates an attention map along the spatial dimension. It uses average-pooling and max-pooling operations along the channel axis to aggregate channel information and then concatenates the two aggregated features.

\begin{align*}
    \mathbf{F}_{avg}^s &\in \mathbb{R}^{1 \times H \times W} = \text{AvgPool}(\mathbf{F}_c') \\
    \mathbf{F}_{max}^s &\in \mathbb{R}^{1 \times H \times W} = \text{MaxPool}(\mathbf{F}_c')
\end{align*}

The concatenated feature is then passed through a convolution layer to produce the spatial attention map.

\begin{align*}
    \mathbf{M}_s(\mathbf{F}_c') &\in \mathbb{R}^{1 \times H \times W} = \sigma(\text{Conv}([\mathbf{F}_{avg}^s; \mathbf{F}_{max}^s])) \\
    \mathbf{F}_s' &\in \mathbb{R}^{C \times H \times W} = \mathbf{M}_s(\mathbf{F}_c') \otimes \mathbf{F}_c' \in \mathbb{R}^{C \times H \times W}
\end{align*}

Here, $[;]$ denotes the concatenation along the channel axis.

\subsubsection{Overall Process}
The overall CBAM process can be summarized as follows:
\begin{enumerate}
    \item Apply the Channel Attention Module to the input feature map $\mathbf{F}$.
    \item Apply the Spatial Attention Module to the refined feature map from the Channel Attention Module.
\end{enumerate}

After applying both attention modules, the refined feature map is the final output $\mathbf{F}_s' \in \mathbb{R}^{C \times H \times W}$.

\subsection{Algorithm}
The following algorithm summarizes the CBAM process:

\begin{algorithm}
\caption{CBAM Attention Module}
\begin{algorithmic}[1]
\State \textbf{Input:} Feature map $\mathbf{F} \in \mathbb{R}^{C \times H \times W}$
\State \textbf{Output:} Refined feature map $\mathbf{F}_s' \in \mathbb{R}^{C \times H \times W}$
\State // Channel Attention Module
\State $\mathbf{F}_{avg}^c \gets \text{AvgPool}(\mathbf{F})$
\State $\mathbf{F}_{max}^c \gets \text{MaxPool}(\mathbf{F})$
\State $\mathbf{M}_c(\mathbf{F}) \gets \sigma(\text{MLP}(\mathbf{F}_{avg}^c) + \text{MLP}(\mathbf{F}_{max}^c))$
\State $\mathbf{F}_c' \gets \mathbf{M}_c(\mathbf{F}) \otimes \mathbf{F}$
\State // Spatial Attention Module
\State $\mathbf{F}_{avg}^s \gets \text{AvgPool}(\mathbf{F}_c')$
\State $\mathbf{F}_{max}^s$ \
\State $\mathbf{F}_{max}^s \gets \text{MaxPool}(\mathbf{F}_c')$
\State $\mathbf{M}_s(\mathbf{F}_c') \gets \sigma(\text{Conv}([\mathbf{F}_{avg}^s; \mathbf{F}_{max}^s]))$
\State $\mathbf{F}_s' \gets \mathbf{M}_s(\mathbf{F}_c') \otimes \mathbf{F}_c'$
\State $\mathbf{F}_{max}^s \gets \text{MaxPool}(\mathbf{F}_c')$
\State $\mathbf{M}_s(\mathbf{F}_c') \gets \sigma(\text{Conv}([\mathbf{F}_{avg}^s; \mathbf{F}_{max}^s]))$
\State $\mathbf{F}_s' \gets \mathbf{M}_s(\mathbf{F}_c') \otimes \mathbf{F}_c'$
\State \textbf{return} $\mathbf{F}_s'$
\end{algorithmic}
\end{algorithm}

\subsection{Squeeze-and-Excitation Network}
The Squeeze-and-Excitation (SENet)~\cite{squeezeexcitation} network introduces a lightweight gating mechanism that can be used to model channel-wise relationships adaptively.

\textbf{SE Block Architecture.}
The SE block can be divided into two main operations: Squeeze and excitation.

\subsubsection{Squeeze Operation}
The squeeze operation aggregates feature maps in spatial dimensions, producing a channel descriptor. For an input feature map $\mathbf{U} \in \mathbb{R}^{H \times W \times C}$, where $H$, $W$, and $C$ are the height, width, and number of channels, respectively, the squeeze operation is defined as:
\begin{equation*}
\mathbf{z}_c = \frac{1}{H \times W} \sum_{i=1}^{H} \sum_{j=1}^{W} \mathbf{U}_{i,j,c}
\end{equation*}
where $\mathbf{z} \in \mathbb{R}^{C}$ is the channel descriptor.

\subsubsection{Excitation Operation}
The excitation operation captures channel-wise dependencies using fully connected (FC) layers. The channel descriptor $\mathbf{z}$ is passed through two FC layers with nonlinearity (ReLU) between. The excitation operation is defined as:
\begin{equation*}
\mathbf{s} = \sigma(\mathbf{W}_2 \delta(\mathbf{W}_1 \mathbf{z}))
\end{equation*}
where $\mathbf{W}_1 \in \mathbb{R}^{\frac{C}{r} \times C}$ and $\mathbf{W}_2 \in \mathbb{R}^{C \times \frac{C}{r}}$ are the weights of the FC layers, $\delta$ denotes the ReLU function, $\sigma$ denotes the sigmoid function, and $r$ is the reduction ratio.

\subsubsection{Recalibration}
The final step is to recalibrate the input feature map $\mathbf{U}$ using the excitation vector $\mathbf{s}$. This is achieved by channel-wise multiplication:
\begin{equation*}
\mathbf{\hat{U}}_c = \mathbf{s}_c \mathbf{U}_c
\end{equation*}
where $\mathbf{\hat{U}} \in \mathbb{R}^{H \times W \times C}$ is the recalibrated feature map.

\subsection{ECA-Net}
ECA-Net~\cite{ecanet} employs a simple but efficient method to capture channel-wise dependencies. The key components of the ECA block are described below:

\subsubsection{Channel Attention Mechanism}
Given an input feature map $\mathbf{U} \in \mathbb{R}^{H \times W \times C}$, the channel attention mechanism in ECA-Net works as follows:

\begin{enumerate}
    \item \textbf{Global Average Pooling:} First, a global average pooling operation is applied to the input feature map to produce a channel-wise descriptor:
    \begin{equation*}
    \mathbf{z}_c = \frac{1}{H \times W} \sum_{i=1}^{H} \sum_{j=1}^{W} \mathbf{U}_{i,j,c}
    \end{equation*}
    where $\mathbf{z} \in \mathbb{R}^{C}$ is the channel descriptor.

    \item \textbf{Efficient Channel Attention:} Instead of using two fully connected (FC) layers, ECA-Net uses a 1D convolution with kernel size $k$ to capture cross-channel interaction. The kernel size $k$ is adaptively determined by a function of channel dimension $C$:
    \begin{equation*}
    k = \psi(C) = \left \lceil \frac{C}{\gamma} \right \rceil \text{, where $\gamma$ is a constant.}
    \end{equation*}
    The efficient channel attention is computed as:
    \begin{equation*}
    \mathbf{s} = \sigma(\text{Conv1D}(\mathbf{z}, k))
    \end{equation*}
    where $\text{Conv1D}$ denotes the 1D convolution operation and $\sigma$ is the activation function of the sigmoid.

    \item \textbf{Recalibration:} The recalibrated feature map is obtained by channel-wise multiplication:
    \begin{equation*}
    \mathbf{\hat{U}}_c = \mathbf{s}_c \mathbf{U}_c
    \end{equation*}
    where $\mathbf{\hat{U}} \in \mathbb{R}^{H \times W \times C}$ is the recalibrated feature map.
\end{enumerate}

\section{Experiments}
This section presents the experimental results of each model using the MSTAR dataset, which has an image size of $(128, 128)$ pixels, $10$ classes, and $2747$ training images followed by $2425$ testing images. We outline the quantitative and qualitative experiments conducted before providing the results for each model and MSTAR. We use MSTAR as our sole dataset; datasets for SAR ATR containing a single target for classification are scarce. Compared to the few publicly available SAR datasets~\cite{opensarship,fusarship}, MSTAR allows us to use the same resolution image on different images, allowing a level comparison across images.

Furthermore, we use ResNet-18 as a base for all self-attention mechanisms, allowing a level comparison between each mechanism. It has been proven to perform well on the MSTAR data set and is convolutional-based, allowing each mechanism we use to seamlessly ``connect'' to the model.

\textbf{Accuracy. }
We run each model in 3 individual tests and calculate each test's average top-1 accuracy metric. 

\textbf{Robustness. }
We perform a robustness test to quantify each model's performance by adding noise. Learning in the presence of noise is essential for  SAR ATR, as many SAR images often contain random backscatter.

\textbf{Explainability. }
We perform a qualitative explainability test by showing the Grad-CAM~\cite{gradcam} saliency map for sample MSTAR images and discussing the implications of each saliency map.

\subsection{Experiments}

\subsubsection{Accuracy}
The top-1 accuracy results for the MSTAR dataset are shown in Table~\ref{tab:mstar}. Notice that each model with added attention slightly outperforms the baseline ResNet-18 model.

\begin{table}[H]
    \centering
    \resizebox{\columnwidth}{!}{
    \begin{tabular}{@{}cccccc@{}}
    \toprule
    Model & Test 1 & Test 2 & Test 3 & Average \\ \midrule
    Standard ResNet-18 & 97.10\% & 97.24\% & 97.45\% & 97.26\% \\ \midrule
    CBAM ResNet-18 & 97.52\% & 97.58\% & 97.89\% & \bf{97.66\% (+0.40\%)} \\ 
    SENet ResNet-18 & 97.24\% & 97.45\% & 97.66\% & 97.45\% (+0.19\%)\\
    ECANet ResNet-18 & 97.17\% & 97.48\% & 97.24\% & 97.30\% (+0.04\%) \\ \bottomrule
    \end{tabular}
    }
    \vspace{1em}
    \caption{Top-1 accuracy of the models on the MSTAR Dataset}
    \label{tab:mstar}
    \end{table}

\subsubsection{Robustness}
We perturb each pixel in each image in the MSTAR dataset with a distribution $\sim \mathcal{N}
\left(0, \frac{3}{255}\right)$ and measure the top-1 accuracy in Table~\ref{tab:robustness}

\begin{table}[H]
    \centering
    \resizebox{\columnwidth}{!}{
    \begin{tabular}{@{}cccccc@{}}
    \toprule
    Model & Test 1 & Test 2 & Test 3 & Average \\ \midrule
    Standard ResNet-18 & 95.39\% & 95.04\% & 95.56\% & 95.33\% \\ \midrule
    CBAM ResNet-18 & 96.33\% & 96.54\% & 97.01\% & \bf{96.63\% (+1.33\%)} \\ 
    SENet ResNet-18 & 96.11\% & 96.09\% & 96.01\% & 96.07\% (+0.74\%)\\
    ECANet ResNet-18 & 95.15\% & 95.57\% & 95.88\% & 95.53\% (+0.20\%) \\ \bottomrule
    \end{tabular}
    }
    \vspace{1em}
    \caption{Top-1 accuracy of the models on the MSTAR Dataset Under a $\sim \mathcal{N}
\left(0, \frac{3}{255}\right)$  Input Perturbation}
    \label{tab:robustness}
    \end{table}

Thus, we confirm that each model is more robust to an input perturbation than the baseline, allowing for a more practical model in the SAR ATR realm, as many SAR images often contain random backscatter.

\subsubsection{Explainability}
\begin{figure}
    \centering
    \includegraphics[scale=0.27]{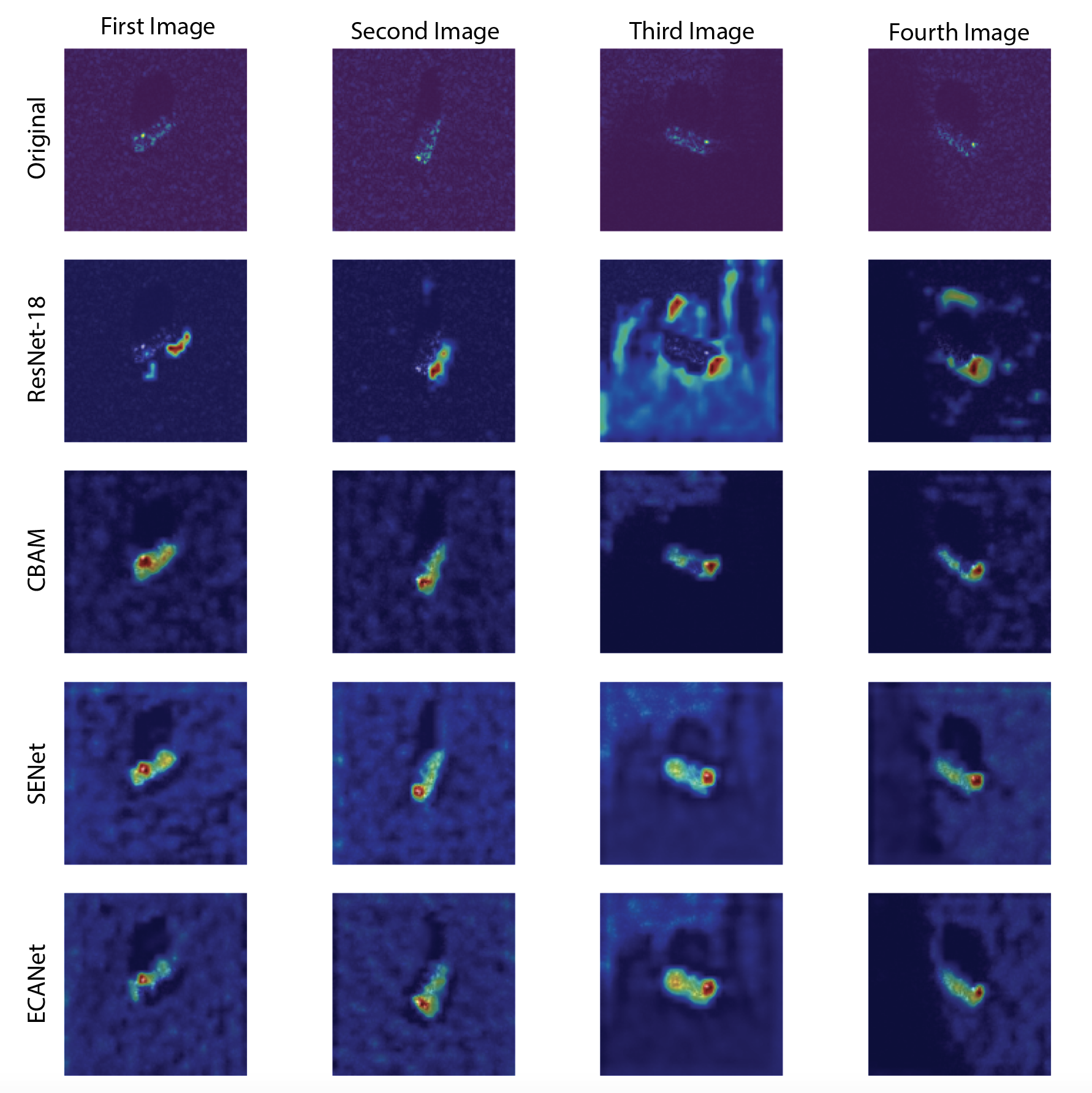}
    \caption{Grad-CAM~\cite{gradcam} activation saliency map from each model with a ResNet backbone on the MSTAR dataset}
    \label{fig:grid}
\end{figure}

We provide the Grad-CAM~\cite{gradcam} activation saliency map for four sample images from the MSTAR dataset in Fig.~\ref{fig:grid}.

\subsection{Discussion}
\subsubsection{Accuracy} Refer to Table~\ref{tab:mstar}. We notice that each model with added attention experiences a slight increase in the top-1 accuracy metric in the quantitative section. The ResNet-18 model~\cite{resnet} with CBAM~\cite{cbam} attention performs the best with an added $0.40\%$ above the baseline in MSTAR.

\subsubsection{Robustness} 
Refer to Table~\ref{tab:robustness}. We notice that each model with self-attention under a perturbation experiences higher top-1 accuracy than the ResNet-18 baseline model.

\subsubsection{Explainability} 
Refer to Fig.~\ref{fig:grid}. The ResNet model randomly learns from areas outside or only parts of the vehicle. Visually, we notice that each attention module added to ResNet-18 can learn from the entire vehicle in each case; however, it also learns slightly from backscatter. Additionally, note that the CBAM module learns the least from the backscatter. 

\subsubsection{General Discussion}
Although the attention modules only add small accuracy benefits for the backbone model we tested, the quantitative portion of this experiment highlights the benefits of including self-attention for SAR ATR. We notice that including self-attention allows each model to learn from the entire vehicle and slightly from the backscatter rather than from random parts of the image, which ResNet seemingly does. Learning in this manner allows for more explainable interactions between practitioners using SAR ATR and deep learning models. We hope that future SAR ATR models consider their design's explainability.

\section{Conclusion and Future Work}
This study evaluated the different mechanisms of self-attention in the computer vision community applied to SAR ATR. Applying self-attention modules for CNN-based architectures allows the model to be guided by vehicle parts rather than random parts of the image, which provides explainable interactions between practitioners and deep learning models. We verified that ResNet-18, as a baseline with added self-attention mechanisms, provides more accurate, robust, and explainable algorithms. Additionally, applying attention modules to SAR ATR algorithms can allow for a more robust model to input perturbations in the presence of noise. Future directions for including self-attention in SAR ATR include domain-specific attention modules for SAR ATR such as~\cite{physicsattention,saratratt}.


\ifCLASSOPTIONcaptionsoff
  \newpage
\fi

\bibliographystyle{IEEEtran}
\bibliography{references}

%








\end{document}